# TRANSLATING SAR TO OPTICAL IMAGES FOR ASSISTED INTERPRETATION

*Shilei Fu, Student Member, IEEE, Feng Xu, Senior Member, IEEE, and Ya-Qiu Jin, Life Fellow, IEEE*
Key Lab for Information Science of Electromagnetic Waves (MoE), Fudan University, Shanghai 200433, China.

**ABSTRACT**
Despite the advantages of all-weather and all-day high-resolution imaging, SAR remote sensing images are much less viewed and used by general people because human vision is not adapted to microwave scattering phenomenon. However, expert interpreters can be trained by compare side-by-side SAR and optical images to learn the translation rules from SAR to optical. This paper attempts to develop machine intelligence that are trainable with large-volume co-registered SAR and optical images to translate SAR image to optical version for assisted SAR interpretation. A novel reciprocal GAN scheme is proposed for this translation task. It is trained and tested on both spaceborne GF-3 and airborne UAVSAR images. Comparisons and analyses are presented for datasets of different resolutions and polarizations. Results show that the proposed translation network works well under many scenarios and it could potentially be used for assisted SAR interpretation.

## 1. INTRODUCTION

Synthetic aperture radar (SAR) is capable of imaging at high resolution in all-day and all-weather conditions. As a cutting-edge technology for space remote sensing, it has found wide applications. Despite the rapid progresses in SAR imaging technologies, the challenge still remains in the interpretation of SAR imagery and it is becoming more and more urgent as a huge volume of SAR data is being acquired daily by numerous radar satellites in orbits.

Due to its distinct imaging mechanism and the complex electromagnetic (EM) wave scattering process, SAR exhibits very different imaging features from optical images. Human's visual system is adapted to the interpretation of optical images. SAR image is difficult to interpret by ordinary people. This has now become the major hindrance in utilization of existing SAR archives and further promotion of SAR applications.

Experts of SAR interpretation are often trained by comparing the SAR image side-by-side with the corresponding optical image. From such SAR-to-Optical comparison experiences, experts conclude useful rules which translate between features in SAR and optical remote sensing images. Thereafter, they are able to direct interpret any new image from similar SAR sensors. Ideally, such training could be done in computers with artificial intelligence (AI). This work is inspired by the recent progresses in AI and deep learning technologies [1-4].

Our major objective is to develop machine learning algorithm which is trainable with large amount of co-registered SAR and optical images to translate from SAR image to optical images and vice versa. The translated optical image can then be used for assisted interpretation of SAR image by massive ordinary people. Imagine that, with such translation tool, any person without any background knowledge of radar, could be able to understand SAR image. This could greatly promote the wide application and usage of future and existing archive of SAR remote sensing imagery. Other potential applications include facilitation of data fusion of optical and SAR images, e.g. translating optical image at an earlier date as the reference SAR image for SAR change detection, registering the unpaired SAR and optical images, integrating hyperspectral SAR into a single image to enhance urban surface features etc.

In this paper, we propose a SAR-Optical image translation generative adversarial network (GAN). A convolutional neural network (CNN) works as the discriminator, while a well-designed network is generator. The generator uses the multi-scale encoder-decoder U-Net as the backbone and incorporates multi-scale residual connections. The GAN is trained alternatively in a min-max fashion. The discriminator is trained to maximize the difference between the co-registered generated sample and true sample, while the generator is trained to minimize it. To reduce the instability during the training of GAN, a hybrid loss function is used to train the generator which contains two parts: the GAN loss back-propagated from discriminator output, and the L1-distance loss directly applied to the generated sample and true sample.

## 2. METHODOLOGY
### 2.1. Network Architecture

The overall framework proposed is shown in Fig. 1. It has two reciprocal directions of translation, i.e. SAR to Optical and Optical to SAR. Each direction consists of two adversary deep networks, i.e. a multi-scale convolutional encoder-and-decoder network as the translator (generator) against a convolutional network as the discriminator. The translator takes in one SAR image, maps it to the latent space via encoder, and then remaps it to a translated optical image. The discriminator takes in both the translated optical image and the true optical image which is co-registered with the original SAR image, and outputs the probability maps. The discriminator learns to identify the translated fake optical image from the true optical images, while the translator network learns to convert the SAR image to an optical image as realistic as possible for the discriminator to fail to separate the two. On the other direction, the network is the same with the only difference being optical as input and SAR as translated image.

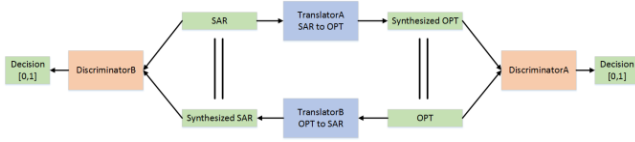

Fig. 1. The conceptual diagram of the translation network.

The discriminator is a conventional CNN for binary classification task. The translator has multi-scale convolutional layers for encoder and decoder where direct paths are connected from the encoder to the decoder at different scales. Beside the direct paths in the latent space, residual connections in the input image space are further incorporated at each scale. A conventional binary classification loss is employed to train the discriminator, while its opposite loss, together with one other losses, are used to train the translator.

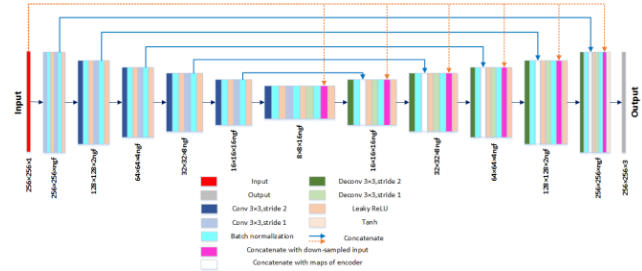

(a) Translator network architecture. The input data size is 256×256×1 and the output data size is 256×256×3. The first two numbers represent the size of the feature maps and the third number represents channels. The concatenation from the encoder and the input to the decoder is signified by lines with arrows. ngf is set as 50.

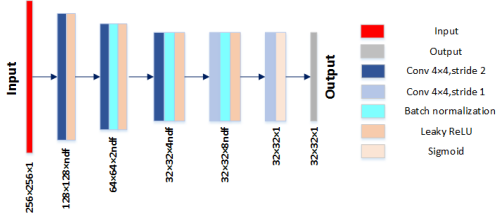

(b) Discriminator network architecture. The input data size is 256×256×1 and the output probability map size is 32×32×1. The first two numbers represent the size of the feature maps and the third number represents channels. ndf is set as 64.

Fig. 2. The architectures of the translation network.

### 2.2. Loss Function

Loss functions are critical for training of the networks. The discriminator [1] is trained with a binary classification log-loss, i.e.

$$L(D) = -E_{x \sim p_{data}(i)}[\log(D(x))] - E_{z \sim p_{data}(j)}[\log(1 - D(T(z)))] \quad (1)$$

where $i = 0,1$ in $p_{data}(i)$ demonstrate the distributions of the true optical and SAR images respectively. $E_{x \sim p_{data}(i)}$ denotes that x obeys to the distribution $p_{data}(i)$, and $E_{z \sim p_{data}(j)}$ denotes that z obeys to the distribution $p_{data}(j)$. When z denotes the original input SAR (or optical) image, $T(z)$ denotes the translated optical (or SAR) image and x denotes the corresponding true optical (or SAR) image. $D(\cdot)$ denotes the output probability map of the discriminator. For the discriminator, minimizing $L(D)$ is equivalent to classifying x as 1 and $T(z)$ as 0.

Following the adversary scheme, the loss function of the translator is

$$L_{GAN}(T) = -\sum_i E_{z \sim p_{data}(i)}[\log(D(T(z)))] \quad (2)$$

where $L_{GAN}(T)$ is the sum loss of the two translated networks. Opposite to the goal of the discriminator, the translator is aimed at synthesizing 'realistic' images to fool the discriminator to classify them as 1.

It is found that the adversary loss function is better to be hybrid with traditional loss, such as L1 [5]. Here, $L_{L1}(T)$ is defined as the L1 distance between the translated image $T(z)$ and the true image x, i.e.

$$L_{L1}(T) = \sum_{i,j} E_{x \sim p_{data}(i), z \sim p_{data}(j)}\left[\|x - T(z)\|_1^1\right] \quad (3)$$

Combine the above two equations together with appropriate weights and derive the final loss function $L(T)$ of the translators.

$$L(T) = L_{GAN}(T) + \beta L_{L1}(T) \quad (4)$$

where $\beta = 20$. $L(T)$ is the objective function for two translators, whose parameters are simultaneously updated. The two discriminators are allocated with the independent loss function $L(D)$ and trained separately.

## 3. EXPERIMENTS

Extensive experiments are conducted to test the performance and generalization capability of the proposed network, including scenarios of high\low-resolution, single\quad-polarization, and different SAR sensors. SAR data used in this study mainly comes from the China spaceborne GF-3 SAR [7] at C-band and the NASA airborne UAVSAR system [6] at L-band. All SAR images are co-registered with a corresponding optical remote sensing image from Google Map™. The original resolution of GF-3 images is 0.51m, while the UAVSAR images are at 6m resolution. For the ease of data handling, images are cut into $256 \times 256$ patches.

## 3.1. Resolution

Fig. 3 compares the translation performance of the network of different SAR data at different resolutions. Note that this is the test dataset, meaning that the network is trained with one set of data and then tested on a different SAR image. It seems that the translated optical image looks very similar to the true image, which is easier for general users to understand. Note that the opposite direction of translating optical to SAR works well too.

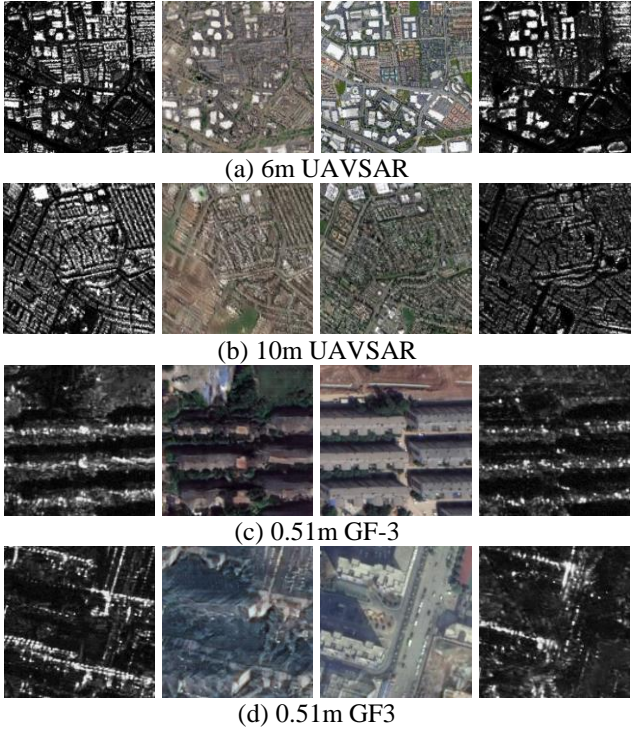

(a) 6m UAVSAR

(b) 10m UAVSAR

(c) 0.51m GF-3

(d) 0.51m GF3

Fig. 3. Translated images with different resolutions. Images at each column from left to right are the original SAR and its translated optical image, the true optical and its translated SAR image, respectively.

Fréchet inception distance (FID) [8] is usually used to quantitatively evaluate the quality and variety of images synthesized by GANs. FID encodes the input image to a probability vector by the inception network, which plays the role of human visual perception system. If the two images are identical, their probability vectors should be the same. FID [8] between the Gaussian distribution with mean and covariance $(m_1, C_1)$ and the Gaussian distribution with $(m_2, C_2)$ is defined as $\| m_1 - m_2 \|_2^2 + Tr(C_1 + C_2 - 2(C_1 C_2)^{1/2})$. Lower FID is better, suggesting more generated samples are similar to true ones. As the dimensionality of the output probability vector is 2048, the number of samples to calculate FID should be greater than 2048 to ensure full rank of the covariance matrix [7]. In our experiment, 2048 pairs of samples are randomly used to calculate the FID. Results are listed in Table 1 where conventional metrics such as the L1 distance, PSNR and SSMI are also calculated for reference.

Table 1 Quantitative metrics of translated images with different resolutions

|      | 0.51m GF3 |         | 6m UAVSAR |         | 10m UAVSAR |         |
|------|-----------|---------|-----------|---------|------------|---------|
|      | optical   | SAR     | optical   | SAR     | optical    | SAR     |
| L1   | 16.6656   | 10.7178 | 6.7340    | 3.7452  | 7.8813     | 12.4393 |
| PSNR | 15.5820   | 15.9172 | 16.1323   | 20.2906 | 16.4238    | 18.3092 |
| SSIM | 0.2799    | 0.2594  | 0.3092    | 0.3640  | 0.3346     | 0.2819  |
| FID  | 154.7532  | 53.0067 | 106.3988  | 56.0202 | 138.3651   | 64.7359 |

## 3.2. Polarization

The UAVSAR data comes with quad-pol information. We wonder how the fully polarimetric information might be useful when translating to optical. We simply use the Pauli pseudo-color SAR image as input and compare it with HH single-pol gray-scale image. An experiment is carried out to train the network with full-pol and single-pol SAR images in the same region, respectively. As shown in Fig. 5, it is found that the optical images translated from full-pol SAR images are more colorful than those from single-pol images.

As explained in Fig. 4, full-pol SAR image apparently contains more information than single-pol. For the same building target (marked in circles), it is much difficult to see in the single-pol then that in the full-pol image. Thus the optical images translated from full-pol images contain more accurate information and appear more vivid. This is also demonstrated with the quantitative metrics listed in Table 2.

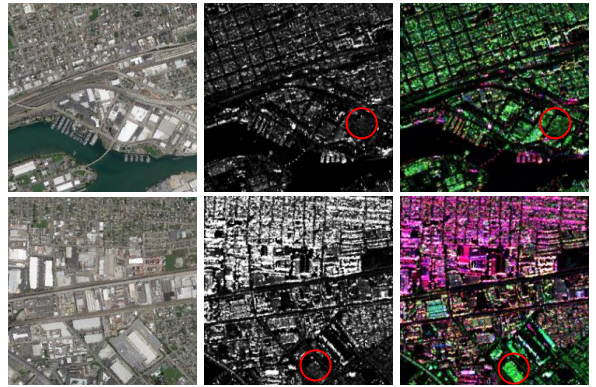

Fig. 4. Comparisons between single-pol and full-pol SAR data. Images in each column: (left) real optical image, (center) real single-polarized SAR image, (right) real full-polarized SAR image.

Table 2 Quantitative metrics of translation images with different polarizations

|      | Single-polarized |         | Full-polarized |         |
|------|------------------|---------|----------------|---------|
|      | optical          | SAR     | optical        | SAR     |
| L1   | 6.7340           | 3.7452  | 8.0453         | 2.7911  |
| PSNR | 16.1323          | 20.2906 | 16.1489        | 19.2188 |
| SSIM | 0.3092           | 0.3640  | 0.3109         | 0.3768  |
| FID  | 106.3988         | 56.0202 | 85.5703        | 52.7645 |

## 4. CONCLUSION

In this paper, a SAR-to-Optical image translation network architecture with reciprocal GANs is proposed as an AI-based assisted interpretation tool for general people to understand SAR images. It consists of a novel multi-scale cascaded decoder-encoder architecture and a hybrid L1 and GAN loss function. In order to evaluate the quality of image translation in the sense of human visual perception, FID is employed as a quantitative measure. For low-resolutions (6m, 10m), the translated optical images appear very similar to the ground truths and the corresponding FID is very good. For high-resolution (0.51m), the translated results appear acceptable but may not accurate capture the geometric features of tall manmade objects especially the high-rise buildings. Comparison is also done with full-pol and single-pol images and it suggests that full-pol SAR image would further improve the translation performance.

Based on this initial attempt, we are currently working towards to train a deep model with large-volume of SAR and optical data and try to improve the generalization capability of the translator network. We expect to put the tool into practical usage.

## ACKNOWLEDGEMENT

This work is supported in part by the National Key R&D Program Grant No. 2017YFB0502703 and the Natural Science Foundation of China Grant No. 61822107.

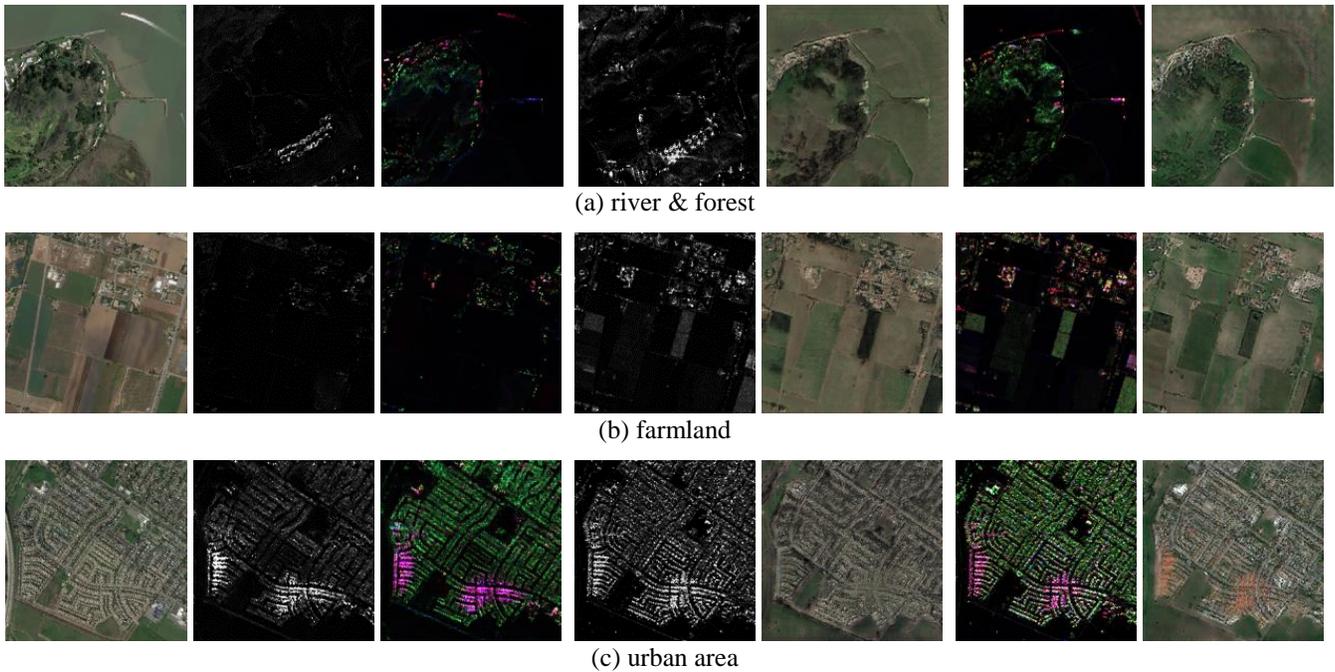

(a) river & forest

(b) farmland

(c) urban area

Fig 5. Images shown above in each column from left to right are: the optical ground truth, and its translated single-pol SAR image and full-pol SAR image, the true single-pol SAR image and its translated optical image, the true full-pol SAR image and its translated optical image. Each row corresponds one kind of earth surface.